%% file: main.tex
\documentclass[11pt, a4paper, logo, twocolumn, copyright]{googledeepmind}

\pdftrailerid{redacted}

\makeatletter
\renewcommand\bibentry[1]{\nocite{#1}{\frenchspacing\@nameuse{BR@r@#1\@extra@b@citeb}}}
\makeatother

\usepackage{kantlipsum, lipsum}
\usepackage{dsfont}
\usepackage{gdm-colors}

\usepackage[]{mdframed}
\usepackage[capitalize,noabbrev]{cleveref}

\usepackage[authoryear, sort&compress, round]{natbib}

\graphicspath{{figures/}}

\title{Scaling laws in wearable human activity recognition}

\correspondingauthor{thoddes@google.com}

\renewcommand{\today}{2025-01-31}

\author[1]{Tom Hoddes}
\author[*, 1]{Alex Bijamov}
\author[*, 1]{Saket Joshi}
\author[*,2,3]{Daniel Roggen}
\author[4,5]{Ali Etemad}
\author[2,6]{Robert Harle}
\author[1]{David Racz}

\affil[*]{Equal contributions}
\affil[1]{Google DeepMind}
\affil[2]{Google LLC}
\affil[3]{School of Engineering \& Informatics, University of Sussex, UK}
\affil[4]{Work done while at Google Research}
\affil[5]{Queen's University, Canada}
\affil[6]{Computer Laboratory, University of Cambridge, UK}

\begin{abstract}
Many  deep architectures and self-supervised pre-training techniques have been proposed for human activity recognition (HAR) from wearable multimodal sensors. 
Scaling laws have the potential to help move towards more principled design by linking model capacity with pre-training data volume.
Yet, scaling laws have not been established for HAR to the same extent as in language and vision. 
By conducting an exhaustive grid search on both amount of pre-training data and Transformer architectures, we establish the first known scaling laws for HAR. 
We show that pre-training loss scales with a power law relationship to amount of data and parameter count and that increasing the number of users in a dataset results in a steeper improvement in performance than increasing data per user, indicating that diversity of pre-training data is important.
We show that these scaling laws translate to downstream performance improvements on three HAR benchmark datasets of postures, modes of locomotion and activities of daily living: UCI HAR and WISDM Phone and WISDM Watch. 
Finally, we suggest some previously published works should be revisited in light of these scaling laws with more adequate model capacities.
\end{abstract}

\begin{document}

\maketitle

\section{Introduction}
\label{sec:introduction}
\input{introduction}

\section{Related Work}
\label{sec:related}
\input{related}

\section{Method}
In this paper we explore scaling laws based on Masked Autoencoder with a ViT encoder for the following reasons:
\begin{itemize}
\item
In language and vision domains, Transformers and Masked Autoencoder have proven to scale well due to high paralellization (e.g. compared to recurrent nets). Furthermore, scaling laws in language and vision were studied using similar architectures, which enables us to draw parallels between domains. 
\item 
\cite{narayanswamy2024scalingwearablefoundationmodels} used this architecture on coarse statistical engineered features. We aim to study scaling laws on the raw sensor signal.
\item 
This architecture has been thoroughly vetted for HAR and is in wide spread use \cite{HARmaskedreconstruction} \cite{freqmae}. By studying scaling laws on this architecture we ensure broader applicability of our findings \cref{sec:conclusion}.
\end{itemize}
\label{sec:method}

\subsection{Scaling Laws}
For our scaling laws, we take a different approach from that in language \cite{hoffmann2022chinchilla} \cite{kaplan2020scalinglawsneurallanguage} and vision \cite{zhai2022scalingvits}. 
In these domains, since data was abundant and compute was the primary constraint, they never completed a full epoch, and thus equated number of steps to amount of data. 
For HAR, however, data is the primary constraint, so we repeat data many times (over 100 epochs) until convergence. We fix the number of pre-training steps to 500,000, which was found experimentally to be sufficient for convergence given our model and data sizes. Since we are able to train even our largest models to convergence, we do not fix the amount of compute. 
Instead, we focus on the capacity of the models (number of parameters). 
This is more directly tied to inference cost than training, which aligns with the priorities of many HAR deployments.

\subsection{Encoder}
\begin{figure}[htb]
\begin{center}
\centerline{\includegraphics[width=1\columnwidth]{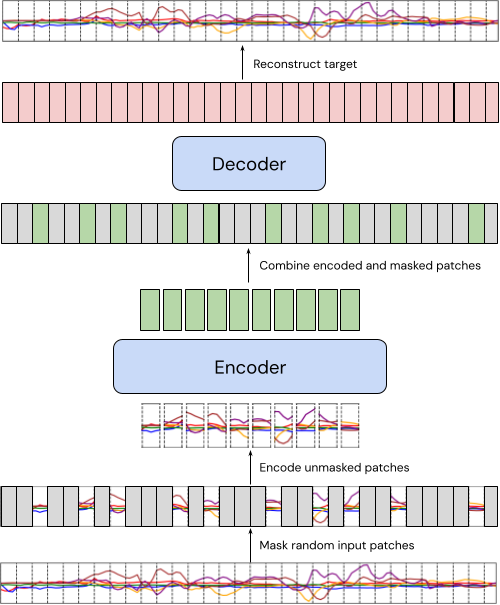}}
\caption{Masked Autoencoder adapted for accelerometer and gyroscope. 
During pre-training, a random subset of accelerometer and gyroscope patches are masked out. Non-masked patches are passed to the encoder and the mask
tokens are re-introduced after the encoder. The encoded patches and mask tokens are then processed by a small decoder trained to reconstruct the original input sequence.
}
\label{mae-architecture}
\end{center}
\end{figure}

For the encoder backbone, we use a ViT \cite{vit} adapted for accelerometer and gyroscope motion sensors as shown in \cref{mae-architecture}. The input to the encoder consists of a time series window of 128 samples at 50Hz, where each sample has 6 channels (x, y, z for accelerometer and gyroscope). We break the window into ``patches" of 4 samples. We choose this patch size to be as small as possible while still fitting the attention matrix in memory. 
Each patch of shape (4, 6) is flattened and transformed linearly to an embedding of dimension size equal to 1/4 the width of the MLP.

We use a standard Transformer block with 8 attention heads. To determine the optimal encoder capacity (i.e. number of parameters) for a given data scale, we conduct a grid search of 3 different widths (512, 1024, 2048 hidden MLP units) and 3 different depths (5, 10, 20 blocks), resulting in 9 different models from 1M to 63M parameters.

\subsection{Pre-training}
Our pre-training approach follows Masked Autoencoder \cite{maskedautoencoder} closely, but adapted for accelerometer and gyroscope motion sensors as shown in \cref{mae-architecture}. This was chosen because it is simple to implement, scales well, and doesn’t require negative examples or domain specific design choices such as augmentations (e.g. to prevent collapse in dual encoders).
We randomly mask whole patches rather than individual samples. 
We only encode unmasked patches and use a high masking ratio of 70\%. This saves considerably on compute costs.
We use a small decoder consisting of 2 Transformer blocks. We apply a linear projection between the encoder and decoder to decrease the width of the decoder by half.

\subsection{Evaluation}
To evaluate pre-trained encoders, we remove the decoder and use global average pooling to attach a linear classification head to the output of the encoder, which we keep frozen. We use linear evaluation as opposed to full fine-tuning to provide a clearer signal of the information extracted from pre-training alone.

\subsection{Datasets}
We use the Extrasensory dataset \cite{extrasensory2017} for pre-training. This dataset contains more than 300k examples of 20 seconds of sensor data from 60 users. Data has been collected while subjects were engaged in regular everyday behavior for several sensors including accelerometer, gyroscope and magnetometer across multiple phone and wearable devices. The dataset is pre-formatted into 5 folds split by user. Each fold is split into a training and a test set. After filtering for missing data we collected 286140 examples, which equates to approximately 1589 hours of data. 
We ignore the activity labels for pre-training. Within each fold, we vary the amount of data by sampling some percentage of examples. 
The USER sampling strategy takes all the data from a randomly selected percentage of the users in the fold.
The RANDOM sampling strategy puts the data of all users in the fold together and draws a random percentage of examples from that.
To control for non-uniform distribution between different sampling strategies and folds, we report results based on the total hours of pre-training data rather than by percent. We only use the phone data, since the watch data does not contain a gyroscope.

For downstream evaluation, we use popular benchmark datasets UCI HAR and WISDM Phone/Watch datasets. We use the full training dataset for all supervised training.
UCI HAR \cite{human_activity_recognition_using_smartphones_240} contains data from 30 volunteers aged 19-48 engaging in 6 modes of locomotion and postures: walking, walking upstairs, walking downstairs, sitting, standing, laying. The accelerometer and gyroscope data is recorded at 50Hz from a smartphone worn on the waist. 
We use the same random partitioning prescribed by the dataset authors (70\% training and 30\% test sets). We also keep the existing raw data preprocessing pipeline, involving noise filtering, 2.56sec sliding windows with 50\% overlap, resulting in 128 samples per window, and use the raw acceleration, not the low-pass filtered version also present in the dataset.

For WISDM we use the 2019 version of the dataset \cite{wisdm_smartphone_and_smartwatch_activity_and_biometrics_dataset__507} comprising 51 subjects performing 18 activities of daily living (postures, locomotion, house chores, nutrition, work-related activities and others) for 3 minutes each. 
We assign the first 2/3 of users to the training set (subjects 1600-1633), and use the remaining 1/3 for evaluation (subjects 1634-1650) similar to previous benchmarks. We split the WISDM dataset into Phone and Watch body positions and evaluate these separately.

We re-sample all datasets to 50Hz and normalize to the same units. None of the datasets contain a null class.

\subsection{Training schedule}
For every model, data combination, we fix the number of steps for pre-training to 500,000 with a batch size of 2048. This equates to over 100 epochs when using 100\% of the data. We use the Adam optimizer with three different learning rates (1e-3, 1e-4, 1e-5) for every model and take the best result. This ensures that each model has sufficient coverage of the parameter search space regardless of size. We apply dropout during pre-training with a rate of 0.1.

\subsection{Compute}
Our exhaustive grid search results in 1620 (3 learning rates * 6 data sizes * 2 sampling strategies * 5 folds * 9 encoder architectures) different hyperparameter combinations for pre-training. Each run takes between 3 and 35 hours to run on 4 TPUv2 chips with larger models running longer. Our total compute used for pre-training is about 62000 TPU-hours.

\begin{table}[htb]
\caption{Best F1 scores of models trained from scratch (FS) vs linear eval (LE) on pre-trained models for each dataset.} 
\label{f1-table}
\begin{center}
\begin{small}
\begin{sc}
\begin{tabular}{lcccr}
\toprule
Data set & FS & LE \\
\midrule
UCI HAR    & 95.1 & 97.9 \\
WISDM Phone    & 31.9 & 34.3 \\
WISDM Watch & 62.6 & 63.1 \\
\bottomrule
\end{tabular}
\end{sc}
\end{small}
\end{center}
\vspace{-1.5em}
\end{table}

\section{Results}

\subsection{Supervised training from scratch baseline}
For each dataset, we conduct a thorough capacity and hyperparameter search of from-scratch models to establish baselines for comparing with pre-trained models. The best results are listed in \cref{f1-table}. We also look at the effect of model capacity on from scratch training. We find that smaller models work better for UCI HAR, with our second smallest model of about 2M parameters performing best. The effect of capacity on WISDM is less clear. It also appears that deeper models perform better than wide models.

\begin{figure}[t]
\begin{center}
\centerline{\includegraphics[width=0.7\columnwidth]{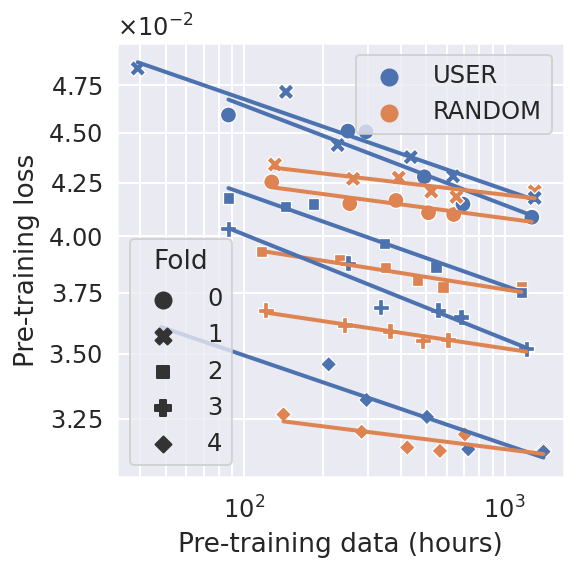}}
\caption{Pre-training test loss vs data size (hours). We fit a power law to each fold and sampling strategy. Equations for each power law can be found in \cref{pretrain-data-law-table}.}
\label{pretrain-loss-data}
\end{center}
\end{figure}

\begin{table}[t]
\caption{Power laws of pre-training test loss vs data size. Exponent values for the USER sampling strategy are roughly 3 times greater than RANDOM.}
\label{pretrain-data-law-table}
\begin{center}
\begin{small}
\begin{sc}
\begin{tabular}{lcccr}
\toprule
Fold & \multicolumn{2}{c}{Sampling Strategy} \\
& USER & RANDOM \\
\midrule
0    & $L = 0.058D^{-0.049}$ & $L = 0.046D^{-0.017}$ \\
1    & $L = 0.057D^{-0.045}$ & $L = 0.047D^{-0.015}$ \\
2    & $L = 0.052D^{-0.046}$ & $L = 0.043D^{-0.020}$ \\
3    & $L = 0.051D^{-0.052}$ & $L = 0.040D^{-0.019}$ \\
4    & $L = 0.043D^{-0.044}$ & $L = 0.035D^{-0.016}$ \\
\bottomrule
\end{tabular}
\end{sc}
\end{small}
\end{center}
\end{table}

\begin{figure}[htb]
\begin{center}
\centerline{\includegraphics[width=0.7\columnwidth]{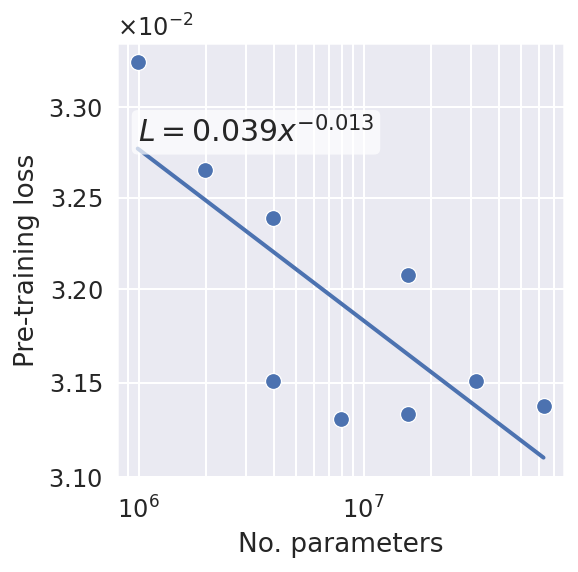}}
\caption{Pre-training test loss vs model capacity (number of parameters) and associated power law fit and equation.}
\label{pretrain-loss-capacity}
\end{center}
\vspace{-1em}
\end{figure}

\begin{figure*}[htb]
\begin{center}
\centerline{\includegraphics[width=0.8\linewidth]{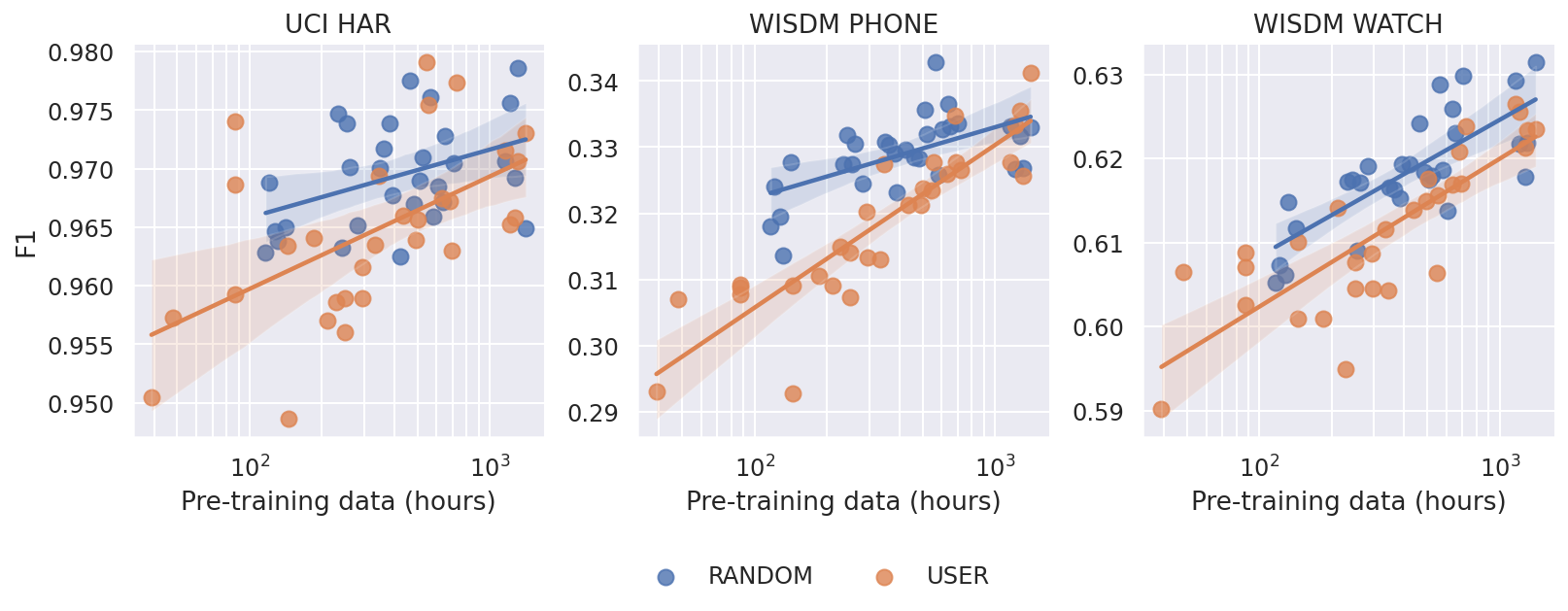}}
\caption{Best linear F1 scores vs pre-training dataset size (hours). Each point represents the best F1 score corresponding to a pre-training fold and data size. The best score is chosen from 27 runs consisting of the 9 encoder architectures and 3 learning rates in our search space.}
\label{pretrain-f1-data}
\end{center}
\vspace{-1em}
\end{figure*}

\begin{figure*}[htb]
\begin{center}
\centerline{\includegraphics[width=0.8\textwidth]{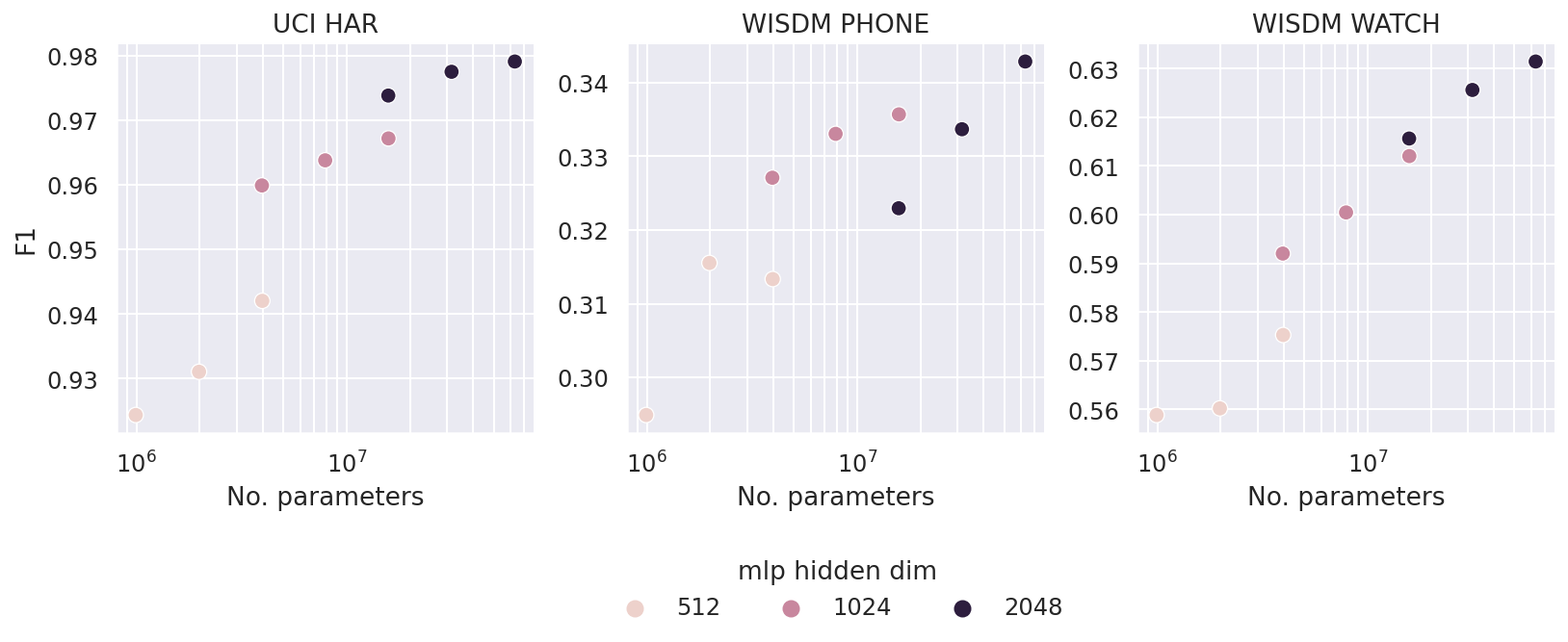}}
\caption{Best linear {F1} scores vs model capacity (number of parameters). Each point represents the best F1 score corresponding to an encoder architecture (width and depth). The best score is chosen from all data sizes and learning rates. We indicate the width (mlp hidden dim) by color. At 5M or 20M parameters we have two models that are the same size, with one wider and shallower (5 blocks) and the other narrower and deeper (20 blocks).}
\label{pretrain-f1-capacity}
\end{center}
\end{figure*}

\subsection{Scaling laws}
\label{sec:result_scaling}

In \cref{pretrain-loss-data} we establish scaling laws of the pre-training test loss vs hours of data. To calculate the loss, we use the full test set from each Extrasensory fold. 
This allows us to compare different training data amounts and distributions within a fold and fit power-law relationships for each fold. 
We observe roughly the same power-law exponent (or slope on the log-log plot) for a given fold and sampling strategy, giving confidence that this relationship was not due to random chance. 
Furthermore, in \cref{pretrain-data-law-table} we see that the exponent is roughly of 3x greater magnitude (or steeper slope) when data is increased by adding more users, as opposed to uniformly or per-user. This emphasizes that diversity of data is extremely important, and dictates the scaling law. Note that the offset is different for each fold, but that is to be expected, since the test sets are different. 
Similarly, in \cref{pretrain-loss-capacity} we fit a power law between pre-training test loss and model capacity in terms of number of parameters, further demonstrating the existence of a scaling law.

\subsection{Downstream performance}
\label{sec:downstream_performance}

We show that the scaling laws for pre-training translate to similar trends in improved downstream linear classification performance. For each pre-training dataset size, we plot the best F1 score from linear evaluation on downstream datasets UCI HAR and WISDM Phone/Watch. This can be seen in \cref{pretrain-f1-data}. Contrary to previous findings \cite{assessingSSLHAR22}, we see consistent improvement as we scale the data size. For UCI HAR, we reach 97.9\% F1 score with linear evaluation. 
To our knowledge, this is on par with the best reported result (98.6\% from \cite{uciharsota}) for this dataset. Note that we do not compare with any public WISDM results because this dataset lacks a common protocol (e.g. train/test split), making direct comparisons irrelevant. 
For all datasets, we surpass from scratch baseline results, with significant improvement for phone datasets UCI HAR (+2.8pp) and WISDM Phone (+2.4pp). Note that WISDM Phone results are low overall, but this can be explained by our protocol including all classes, some of which are difficult or impossible to detect from phone alone (e.g. typing).
For WISDM Watch, the improvement is smaller (+0.5pp). This is not surprising given that our pre-training dataset consists of only phone data. Still, the consistent increase in watch performance suggests that we are seeing positive transfer between body positions.

\begin{figure}[htb]
\begin{center}
\centerline{\includegraphics[width=\columnwidth]{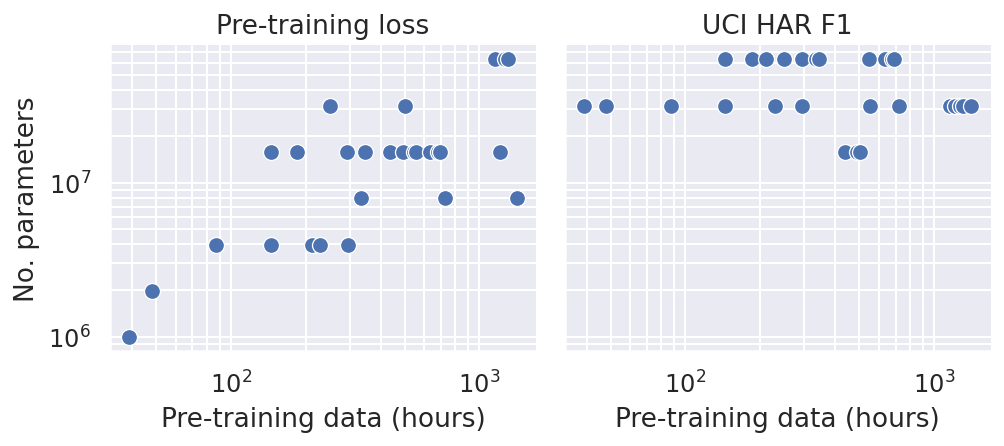}}
\caption{Optimal capacity for a given pre-training data size. Each plot shows the parameter count of the model resulting in the best performance for a given metric (pre-train test loss on the left, UCI HAR test F1 on the right).}
\label{capacity-vs-data}
\end{center}
\vspace{-1em}
\end{figure}

In \cref{pretrain-f1-capacity} we study the effect the capacity of the encoder has on downstream performance. For each of the 9 encoder architectures (3 widths by 3 depths), we plot the best F1 score out of 180 runs (5 folds * 6 data sizes * 2 sampling strategies * 3 learning rates) from linear evaluation on downstream datasets UCI HAR and WISDM vs the number of parameters. We find that increasing the number of parameters is crucial to realizing performance improvements across all 3 tasks. The optimal capacity is reached at our biggest model which has about 63M parameters. This is in contrast to our from scratch baselines, where performance can peak at smaller models (e.g. about 2M parameters for UCI HAR).

\begin{figure*}[t]
\begin{center}
\centerline{\includegraphics[width=0.8\textwidth]{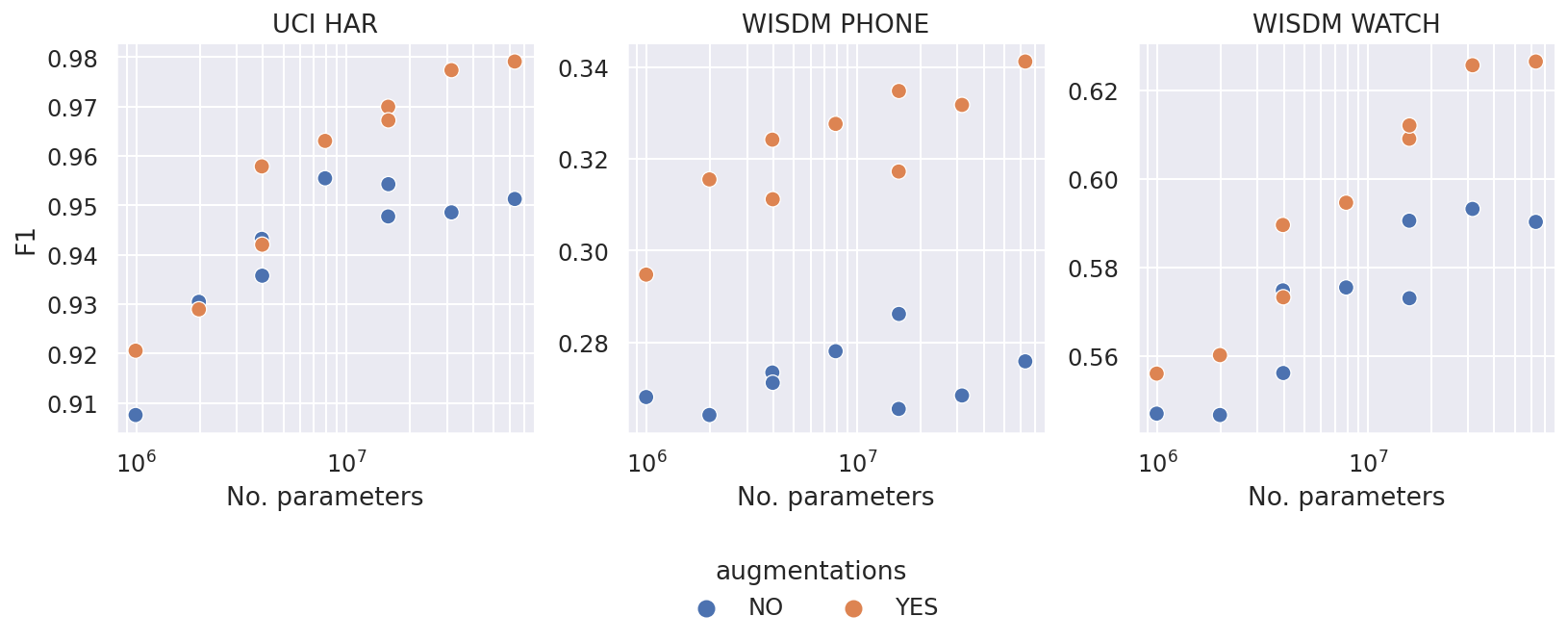}}
\caption{Best linear {F1} scores vs model capacity (number of parameters). Each point represents the best F1 score corresponding to an encoder architecture (width and depth) and whether augmentations were on or off. The best score is chosen from all data sizes and learning rates.}
\label{pretrain-augmentations}
\end{center}
\vspace{-1em}
\end{figure*}

\subsection{Optimal capacity vs data size}
In \cref{capacity-vs-data} we study the optimal capacity for a given pre-training data size. For pre-training test loss, we find that optimal model size increases monotonically with more data. Downstream F1 performance tells a different story, with our largest models performing best even with minimal data. We hypothesize that this may be due to epoch-wise double descent \cite{double-descent} behaviour, which we have observed in some cases in this work.

\subsection{Augmentations}
\label{sec:augmentations}
We apply random rotation and scaling augmentations during pre-training, and study the effect on downstream model performance. In \cref{pretrain-augmentations} We separate results by encoder as in \cref{pretrain-f1-capacity}, but take the best score both with and without augmentations. We see that augmentations always improve performance, especially at larger scales. The optimal capacity without augmentations can be smaller, at either the 4th largest model (about 10M parameters) for UCI HAR or the 3rd largest model (20M parameters) for WISDM Phone. This is not surprising, since augmentations can be thought of as a strong regularizer, and/or an artificial expansion of the dataset.

\subsection{Width vs Depth}
Since we conducted a grid search on width and depth, we have results for a variety of width to depth ratios. There are also 2 model sizes (20M and 5M) for which we have a very deep model (20 blocks) and a very wide model (5 blocks) with the same number of parameters. This allows us to control for the total number of parameters. Looking at \cref{pretrain-f1-capacity}, we can see that increasing both width and depth improve performance, but wider models tend to perform better than deeper models for the same number of parameters.

\section{Conclusion}
\label{sec:conclusion}
\input{conclusion}

\bibliographystyle{abbrvnat}
\nobibliography*
\bibliography{main}

\end{document}

%% file: introduction.tex
Wearable human activity recognition (HAR) aims to recognize the actions of people 
from the time series data originating from the sensors in their wearable devices and mobile phones. These sensors are typically motion sensors, such as triaxial accelerometers, gyroscopes, magnetometers, or their combination in inertial measurement units (IMU). 
This is an important area of research in mobile and ubiquitous computing \cite{Cook10,SanSegundo18,Ploetz18}.
It has applications in domains such as industrial assistance
\cite{Scholl15},
personalized healthcare 
\cite{Lee18}, 
human-computer interaction \cite{Luckowicz10_computer}, 
or sports tracking 
\cite{Hsu19}, well illustrated in many current smartwatches automatically detect fitness workouts.

A wide range of architectures (e.g., convolutional, recurrent, Transformers \textemdash see \cref{sec:related}) have been suggested \cite{Chen21,Gu21}, but so far design principles to create architectures for specific recognition problems and trade-offs still elude us.
This is particularly important as HAR has unique characteristics, distinct from other time series problem (e.g., speech and auditory scene analysis) \cite{Demrozi20}. Notably, the problem is often tackled with several multimodal sensors (e.g., smartphone, smartwatch and physiological sensors). The sensor sample rate can differ by orders of magnitude in a same system\footnote{For example, sound used to detect sound-generating activities (e.g., washing hands, use of a microwave) is sampled in the tens of KHz. Motion and physiological sensors tend to be sampled in the tens to hundreds of Hz. Other sensors may be sampled at a much lower rate, such as GPS sampled in the order of a few Hz.}. The activities themselves can differ wildly in duration\footnote{For example, sporadic events such as ``taking a sip from a glass'' to longer activities such as workouts in a gym, or even longer daily routines}. New sensors are also continuously developed, such as sensorized textiles \cite{TellezVillamizar24}, and power considerations are important when models are to run on embedded devices.
This variety leads to application-specific annotated datasets for training, which are costly to acquire \cite{Welbourne14}, effectively leading to annotation scarcity \cite{Chen21} and growing interest in self-supervised pre-training to address this \cite{SSLHARsurvey}. 
It also makes it challenging to come up with generalizable architectures.
The architectures reported in the literature are generally a result of experimentation and heuristics tailored to application-specific trade-offs.

A promising approach to achieve a more principled design include relying on
scaling laws, which aim to link model capacity, data volume and performance. Scaling laws have been explored in language foundation models \cite{hoffmann2022chinchilla} but there has been little work verifying their existence and benefits in wearable HAR.

Identifying whether such scaling laws exist has important benefits.
Firstly, 
if compute resources are not a limitation, then it is important to design a model with a capacity commensurate to the amount of pre-training data to make best use of it. A model without enough capacity would not be able to take full benefit from the pre-training data. In fact, previous work stated that self-supervised learning data volume is only useful up to a point \cite{assessingSSLHAR22}. Our paper makes the point that in that instance, the model capacity was insufficient to make use of larger volume of data.
A model with too much capacity might further improve classification performance, but it would do so at higher compute cost which has its own downsides (e.g., on embedded devices).

Secondly, 
if compute resources are limited (e.g., constraints to run on embedded hardware, or financial training costs), then there is a corresponding suitable volume of pre-training data. 
Therefore it becomes important to use the most suitable, higher quality, pre-training data available first, given the data volume budget, and only include other lesser quality data if the data volume budget allows for it. 
This means favoring pre-training datasets similar to the target domain: similar location of the sensors (on body placement {\em and} orientation), same modalities (a gyroscope cannot be substituted by an accelerometer), similar sensing characteristics (sample rate, dynamic range), 
similar user activities (static or slow-moving yoga poses may not be suitable to pre-train a model for highly dynamic activities such as contact sports)\footnote{Pre-training on data that from a different domain still works, but this becomes a form of regularization, rather than the more desirable ability to hierarchically capture dynamic and cross-modal properties of the sensor signals.}.
Thus, such scaling laws may provide guidance on how to allocate costs across compute costs for model training and engineering effort for data curation\footnote{Unlabelled data is {\em more easily} available than annotated data, but it does not come ``for free'': it still needs to be identified and curated. This is especially the case in wearable computing where it is difficult to find large pre-existing datasets which may match the specific characteristics of a new recognition problem.
}. 

In short, the research questions of this paper are:

\begin{mdframed}

\noindent {\bf RQ1: } Do scaling laws linking data, model capacity, and performance exist in the data and inference compute constrained domain of HAR?

\noindent {\bf RQ2: } How does data diversity affect the scaling laws?

\noindent {\bf RQ3: } Do these scaling laws suggest new experiments or improvements for previously published work (e.g. models proposed so far having insufficient capacity)?
\end{mdframed}

This paper investigates scaling laws linking pre-training data volume, model capacity and performance, in the context of a ViT adapted for HAR pre-trained on unlabelled data.
Our contributions are:
\begin{itemize}
\item 
Using the large scale public Extrasensory dataset of activities of daily living collected in the wild from smartphone sensors (5000 hours of data from 60 users), we identify for the first time scaling laws in HAR showing that pre-training loss scales with a power law relationship to amount of data and parameter count (\cref{sec:result_scaling}), addressing \textbf{RQ1}.
\item 
We verify that these laws inform the performance on downstream HAR tasks, using 3 datasets of postures, modes of locomotion and activities of daily living, including up to 18 activity classes: UCI HAR and WISDM Phone and WISDM Watch (\cref{sec:downstream_performance}).
\item 
We show that model capacity can be further increased with datasets that are augmented by signal transformations and therefore help alleviate annotation scarcity (\cref{sec:augmentations}).
\item 
We indicate how these scaling laws can be used to inform future research.
Notably, we discuss how these scaling laws can be used to re-interpret previously published work, addressing \textbf{RQ3}. We give two examples where we assess that increased model capacity could have been beneficial (\cref{sec:conclusion}).
Our findings reiterate the importance of pre-training with greater ``diversity'' data rather than solely larger volume of data: we show that adding more users to the pre-training dataset is better than adding more data from the same user.
\end{itemize}

%% file: related.tex
\subsection{Deep learning for wearable activity recognition}
{DeepConvLSTM} was one of the first deep models to outperform classical machine learning pipelines on a benchmark datasets of activities of daily living. It consisted of 4 deep convolutional layers and 2 LSTM layers with a total 3.97M parameters \cite{Ordonez16a}.
As multimodality is important in HAR \cite{Munzner17} explored different fusion strategies based on convolutional architectures, with the largest model containing 7M parameters.
Since then, new models have been suggested with GRU units, attention mechanism, various normalization strategies, reflecting advances in other {ML} fields. 
One oft-cited architecture is Attend and Discriminate, which aims to leverage cross-modal sensor interaction using spatial attention mechanism on top of convolutional layers, and substitutes LSTM units by GRU \cite{Abedin20} (parameter count not reported).
More exhaustive reviews can be found in \cite{Chen21,Gu21}. 

Designing HAR models tends to follow engineering heuristic and design guidelines are still lacking.  Besides scaling laws mentioned in Sec. \ref{sec:introduction}, neural architecture search has been proposed to systematically explore the design space for a particular recognition problem \cite{Wang21}. \cite{Pellatt22} reports larger models than DeepConvLSTM, although the results are reported in FLOPS rather than parameter count (47.2M FLOPS compared to 5.3M for DeepConvLSTM).

Researchers in the field of mobile and wearable computing have tended to minimize compute cost rather than scaling up models in order to embed these models in battery-operated devices. 
TinyHAR performs multi-modal fusion with spatial and temporal Transformer blocks with the largest model having 165K parameters and also outperforming DeepConvLSTM \cite{Zhou22}. Similarly, a shallower version of DeepConvLSTM was proposed in \cite{Bock21} with 63\% less parameters and similar performance. 
Device constraints may seem to contradict the premises of this paper aiming to explore scaling laws and large models. However there are arguments for scaled up models when recognition performance is paramount: 1) pervasive network connectivity allows models to be running in the cloud; 2) scaled-up models can be distilled and quantized to smaller sizes customized for inference on a variety of devices.

\subsection{Self-supervised learning for HAR}

Annotation scarcity in activity recognition can be combated through self-supervised learning, where a pretext task is learnt on an unannotated dataset, with fine-tuning on a smaller annotated dataset.
An exhaustive review of methods can be found in \cite{assessingSSLHAR22,SSLHARsurvey}.
These reviews highlight that variations of masked reconstruction are commonly used.
\cite{HARmaskedreconstruction} proposes a Transformer encoder in the time domain, yielding a model with 1.5M parameters. They only mask 10\% of data and use an MLP decoder. 
SelfPAB \cite{selfpab} \cite{monoselfpab} and FreqMAE \cite{freqmae} use masked autoencoder pre-trained on the spectrograms of the input. 
SelfPAB is inspired by audio models, and does pre-training on the HUNT4 dataset with 100k hours of sensor signals.
FreqMAE is suggesting a specific Transformer block to account for spectral properties of the input signal. Our approach differs from these by faithfully applying masked autoencoder \cite{maskedautoencoder} with minimal change from the vision domain.

\subsection{Scaling laws in language, vision and HAR}

Scaling up the size of Transformers has led to significant improvements in performance in language and vision models.
We are particularly interested in the scaling {\em laws} that link model capacity, pre-training data volume and performance, as this contributes to more principled design.

In language foundation models, \cite{kaplan2020scalinglawsneurallanguage} demonstrated that performance improves as pre-training data and model capacity is increased with a power law relationship. \cite{hoffmann2022chinchilla} built on this and demonstrated that data and model capacity should be scaled equally.

Vision Transformers with masked pre-training \cite{vit} have been shown to perform increasingly better as the model size increased (from 86M to 632M parameters), later even up to 22B parameters \cite{Dehghani23a}, also verified in \cite{maskedautoencoder}.
A saturating power-law linking performance, data and compute was presented in \cite{zhai2022scalingvits}, similarly to language.

Work exploring scaling laws in HAR is less established but there is also evidence that more pre-training data is beneficial. 
\cite{SSLUKbiobank} exploited the 700k hour UK Biobank dataset and showed this on a ResNet encoder with 10M parameter.
These benefits are also reported in \cite{assessingSSLHAR22}, but only to a point. 
\cite{howmuchdataSSLHAR23} tried to identify what is the minimum amount of pre-training data which is required, after which a plateau is reached. The authors clarify their intent is not to identify scaling laws, but rather that identifying {\em ``minimal quantities can be of great importance as they can result in substantial savings during pre-training as well
as inform the data collection protocols''}.
None of these work draw explicit scaling laws.
SelfPAB \cite{selfpab} \cite{monoselfpab} varied data and model capacities and scale Transformers to a size of 60M parameters, similar to this present work. However, our work differs from theirs by establishing scaling laws that link model capacity, data, and performance.

Scaling laws have been explored in sensor foundation models \cite{narayanswamy2024scalingwearablefoundationmodels,Shapiro24}. 
The work in \cite{narayanswamy2024scalingwearablefoundationmodels} differs fundamentally from ours.
The authors create a foundational model not on raw motion sensor data but on a set of 10 engineered statistical features extracted from the motion sensors at a rate of one vector every minute, as well as physiological sensors. 
Although they introduce scaling laws, operating on engineered features makes it difficult to draw direct parallels to our work.
Furthermore, they rely on a proprietary dataset, and report scaling in terms of reconstruction performance (mean squared error) instead of downstream classification performance. 
Our work instead uses public datasets to help reproducibility, and we introduce new conclusions regarding data diversity (\cref{sec:conclusion}).
\cite{Shapiro24}  mention scaling up models as a future research direction and do not yet draw scaling laws from their experiments.

%% file: conclusion.tex
We establish the first known scaling laws for HAR Transformer models and validate their performance on 3 downstream HAR datasets (UCI HAR, WISDM Phone and WISDM Watch), with a large number of activity classes (18 in WISDM). We show that performance improves with a power-law relationship as data is increased, and that the parameters of the power-law depend directly on the diversity of the data being added. 
For example, we find that adding new users results in a power-law exponent that is 3x larger than adding more data from the same users.

Our evaluations differ from those in language and vision  by training to convergence in order to study the relationship between number of parameters and data under the unique conditions of HAR, where we are more constrained by data and inference compute than training compute.

We link model capacity directly to pre-training data by showing that larger models are required in order to take advantage of more pre-training data. In fact, even in the low data regime, we see evidence that it's often worth exploring over-parameterized Transformers. In short: when in doubt, increase the model capacity and train for longer, assuming sufficient resources.

We recommend revisiting previous works that may be under-parameterized. For example \cite{SSLUKbiobank} used the large UK Biobank dataset but fixed the encoder architecture with a 10M parameter ResNet. Masked Reconstruction \cite{HARmaskedreconstruction} was explored in \cite{assessingSSLHAR22} using the large Capture-24 \cite{capture24} dataset, but they fixed the Transformer encoder architecture at 1.5M parameters. We would expect these works to benefit from increasing the capacity to at least 30M parameters.

Even our largest model fits on a single TPUv2, and completes pre-training in under 12 TPU-days. Given this and our focus on public datasets, we believe our results are possible to replicate.
As we scale to larger models, it becomes less feasible to deploy these directly on-device. We recommend using large, pre-trained models as teachers that can be distilled and quantized to smaller, on-device student models.

Future work could verify further the existence of these scaling laws in other previously proposed models, such as \cite{Abedin20}, \cite{selfpab} and contrastive pre-training techniques such as \cite{simclr}. While our results were obtained from wearable motion sensor data, future work could also verify their existence when other sensor modalities are used, such as radar or WiFi which become increasingly more frequently explored in the field.